\documentclass[letterpaper,10 pt,conference]{ieeeconf}  

\IEEEoverridecommandlockouts                              

\usepackage[linesnumbered,ruled,vlined]{algorithm2e}
\pdfoutput=1
\usepackage{graphicx}

\usepackage[caption=false,font=footnotesize,lofdepth,lotdepth]{subfig}
\usepackage{float}
\usepackage{graphicx,adjustbox}
\usepackage{cite}
\usepackage{cancel}
\usepackage{multicol}
\usepackage{array, multirow}
\usepackage{tabulary}
\usepackage{tikz}
\usetikzlibrary{positioning}
\usetikzlibrary{shapes.geometric}
\usetikzlibrary{shapes.misc}
\usepackage{changes}
\renewcommand{\baselinestretch}{0.981}

\title{\LARGE \bf
Cautious Planning with {\disharevised Incremental Symbolic} Perception:\\ Designing Verified Reactive Driving Maneuvers
}

\author{Disha Kamale$^1$, Sofie Haesaert$^2$, Cristian-Ioan Vasile$^1$
\thanks{$^{1}$Disha Kamale and Cristian-Ioan Vasile are with the Mechanical Engineering and Mechanics Department,
Lehigh University, Bethlehem, PA 18015
{\tt\small \{ddk320, cvasile\}@lehigh.edu}}%
\thanks{$^{2}$ Sofie Haesaert is with the Department of Electrical Engineering,
Eindhoven University of Technology, Eindhoven, Netherlands
{\tt\small S.Haesaert@tue.nl}}%
}

\usepackage{graphicx}
\usepackage{amsmath}
\usepackage{amssymb}
\usepackage{amsfonts}
\usepackage{mathtools}
\usepackage{tikz}
\usepackage{xcolor}
\usepackage{amsmath}
\usepackage{amsthm}

\newtheorem{definition}{Definition}

\newcommand{\red}[1]{{\color{red}#1}}
\newcommand{\blue}[1]{{\color{blue}#1}}

\newcommand{\True}{\mathsf{true}}
\newcommand{\disha}[1]{\color{black}#1}

\newcommand{\disharevised}[1]{\color{black}#1}
\newcommand{\dishalatest}[1]{\color{black}#1}
\newcommand{\Next}{\bigcirc}
\newcommand{\Until}{{\mathbin{\sf U}}}
\newcommand{\word}{\pi}
\newcommand{\letter}{l}

		\newcommand{\always}{\square}
\newcommand{\nex}{\bigcirc}
 \newcommand{\eventuallyi}[1]{\lozenge^{#1}}   
 \newcommand{\eventually}{\lozenge}
\newcommand{\until}{{\mathbin{\sf U}}}
\newcommand{\release}{{\mathbin{\sf R}}}
 \newcommand{\tru}{\mathsf{true}}

\newcommand{\meal}{{\mc M}}
\renewcommand{\l}{\left}
\renewcommand{\r}{\right}

\newcommand{\ft}{\footnotesize}
\newcommand{\mc}[1]{{\mathcal{#1}}}
\newcommand{\mb}[1]{{\mathbf{#1}}}
\newcommand{\mbb}[1]{{\mathbb{#1}}}
\newcommand{\mtt}[1]{\mbox{\texttt{#1}}}

\newcommand{\al}[1]{{\mathcal{#1}}  }  

\newcommand{\Env}{{\mc X}}
\newcommand{\Sys}{{\mc Y}}
\newcommand{\Mem}{{\mc E}}
\newcommand{\Dom}{{\Gamma}}
\newcommand{\sx}{a}
\newcommand{\labels}{{\mc Z}}
\newcommand{\Act}{{Act}}

\newcommand{\lab}[1]{\mbox{``#1"}}

\newcommand{\tab}{\mbox{ list }}
\newcommand{\sys}{\mbox{ sys }}
\newcommand{\eps}{\varepsilon}
\newcommand{\apin}{\alpha}
\newcommand{\apout}{\beta}
\newcommand{\Sigin}{\Sigma_{I}}
\newcommand{\Sigout}{\Sigma_{O}}

\newcommand{\AP}{{AP}}
\newcommand{\Lab}{{ \mathrm{Lab}}}   
\newcommand{\Labex}{L}      
\newcommand{\Coloneqq}{::=}      %
\newcommand{\alphabeth}{{\Sigma}}
\newcommand{\ap}{{p_i}}            
\newcommand{\alwaysi}[1]{\Box^{#1}}  
\newtheorem{problem}{Problem}[section]

\newcommand{\TS}{\mathcal{T}}

\begin{document}

\maketitle
\thispagestyle{empty}
\pagestyle{empty}

\begin{abstract}
This work presents a step towards utilizing incrementally-improving symbolic perception knowledge of the robot's surroundings for provably correct reactive control synthesis applied to an autonomous driving problem.
Combining abstract models of motion control and information gathering, we show that assume-guarantee specifications (a subclass of Linear Temporal Logic) can be used to define and resolve traffic rules for cautious planning. \textcolor{black}{We propose a novel representation called }
\textit{symbolic refinement tree for perception}
	    \textcolor{black}{that  captures} the incremental knowledge about the environment 
     and embodies the relationships between various symbolic perception inputs.
		The incremental knowledge is leveraged for synthesizing {\disharevised verified reactive} plans for the robot. The case studies demonstrate the efficacy of the proposed approach in synthesizing control inputs even in case of partially occluded environments. 
\end{abstract}
\section{Introduction}
\label{sec:intro}


Autonomous robots performing highly complex tasks such as urban driving may lead to serious repercussions when malfunctioning. Thus, having an autonomous system
that not only adapts to the changing environment \disharevised {by utilizing the information gathered at runtime (\textit{reactive})} but
also guarantees \textit{safe} behavior becomes crucial \cite{fremont2021safety}. 

\begin{figure}[t]
	\centering
	\includegraphics[width=.97\columnwidth, trim= 0cm 0cm 0cm 0cm, clip]{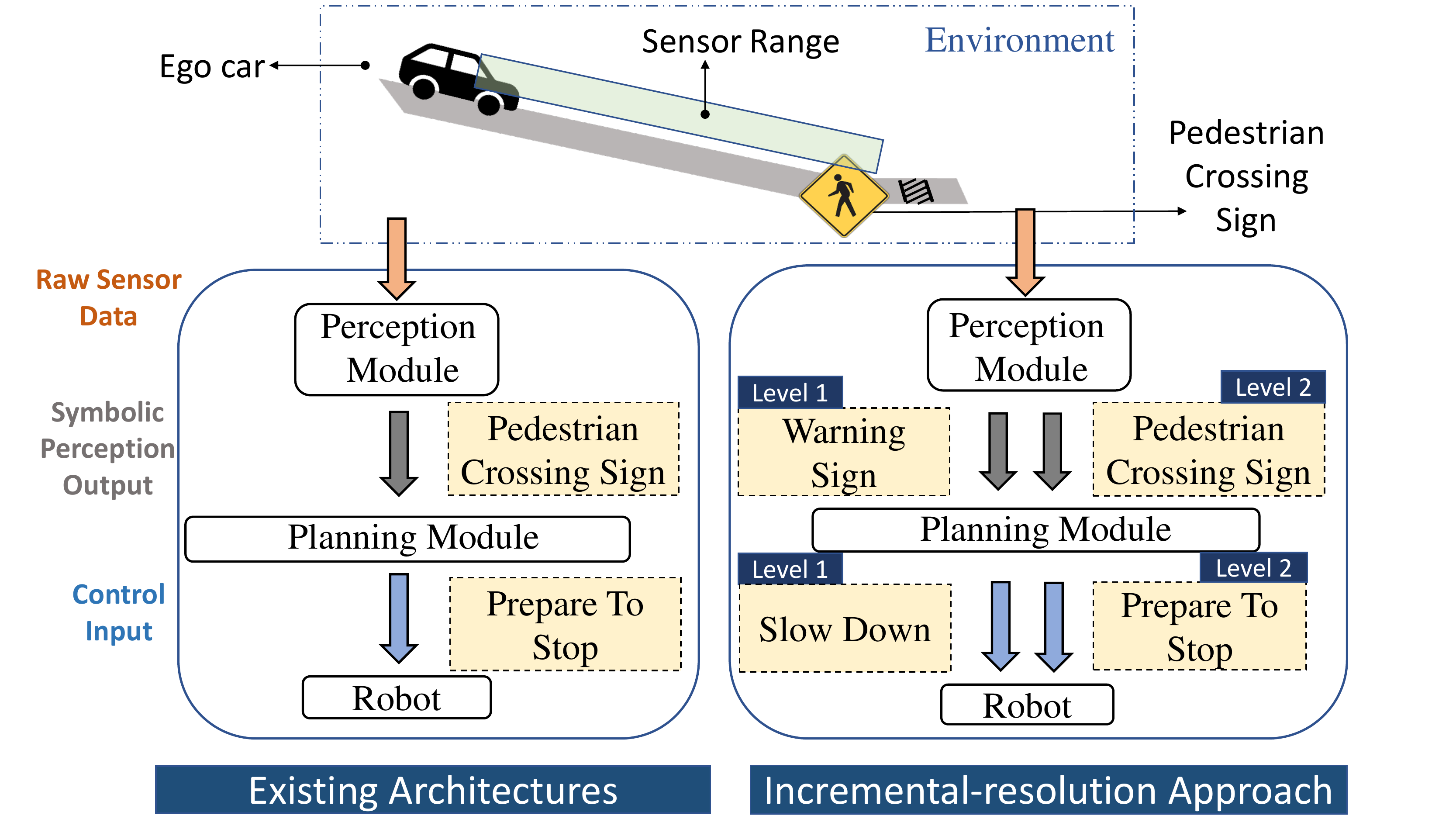}
	\caption{\small{{\disharevised Difference between existing approaches and the proposed incremental-resolution architecture. The green rectangle depicts the range of the onboard sensors. The colored arrows indicate the flow of information as follows. Red: raw sensor input to the perception module, Grey: symbolic perception information as input to planning module, Blue: control input to robot.}}}
 	\label{fig:motivation}
	\vspace{-10pt}
\end{figure}

\disharevised{
This broad problem is usually addressed using modular perception-planning frameworks (e.g.,~\cite{okumura2016challenges, wang2018perception}). Within the perception module, it is often assumed that the targets (objects of interest) can be fully identified assuming, inter alia, perfect sensing, fully observable environment, accurate detection and recognition frameworks, etc., \cite{fayyad2020deep, wang2018perception, fainekos_temporal_2005,finucane_ltlmop_2010,kress-gazit_wheres_2007,sahin_multirobot_2019,wang_hyperproperties_2020}.
However, pertaining to hardware limitations, occlusions, weather conditions, etc., the perception is often limited. Thus, the environment information may not be gathered all at once but may instead incrementally improve as the distance {\dishalatest{to}} target decreases. This renders the classical perception-planning architectures unable to react in a timely manner in cases where the robot may only have {\dishalatest{sparse}} perception information at a given instant. For instance, the autonomous car traveling downhill (Fig.~\ref{fig:motivation}) may detect a yellow board implying a warning sign a few instances before recognizing the exact sign.
If the control module only acts based on the exact knowledge of upcoming environment states (e.g., \cite{fainekos_temporal_2005,finucane_ltlmop_2010, kress-gazit_wheres_2007,sahin_multirobot_2019,wang_hyperproperties_2020}), it may lead to an overly conservative or unnecessarily risky behavior. Such challenging situations give rise to the question: how can partial knowledge be utilized to ensure safe planning? }

{\disharevised The overarching goal of this work is to address the problem of symbolic perception-aware planning where the perception improves as the robot physically progresses towards a target object. To achieve the verifiable usage of perception in control, we use abstractions to encapsulate the detection and estimation techniques (e.g., learning-based approaches, filtering) {\dishalatest{and subsequently}} 
assume symbolic 
information as input which is a commonly followed approach e.g., \cite{vasile2020reactive, ulusoy2014receding, kress2009temporal, kress2011mitigating, liu2013synthesis, vasile2013sampling, kamale2021automata, haesaert2019temporal}. We propose an incremental-resolution representation for perception as a directed tree where each level corresponds to a symbolic feature identified at a given instant.} 
To the authors' best knowledge, this is the first work that utilizes the evolution of perception knowledge in the form of incremental symbolic updates while providing guarantees on the system behavior for possibly adversarial environments. 
\smallskip

\noindent{\textit{A. Symbolic Perception: }}\\
Our approach for symbolic refinement tree representation is inspired by the way in which the human brain abstracts varying levels of details in the surroundings to guide learning and decision-making e.g.,  \cite{palmer1977hierarchical, eckstein2020computational}.
With representation learning as the primary objective, \cite{galindo2005multi} proposes learning hierarchies for obtaining semantically similar regions whereas \cite{konidaris2016constructing} explores learning abstract representations for perception coupled with agent's skills. Instead, our work prescribes symbolic representation and utilizes it for safe control synthesis; learning the relationships among symbolic perception updates is a topic for future research. 

Some existing sensor fusion approaches for perception of autonomous vehicles consider varying levels of detection and processing of individual sensor information~\cite{fayyad2020deep}. Specific to traffic signs, \cite{fan2021multi}, \cite{liang2020traffic} propose the use of feature pyramids for neural networks that capture the low and high-level semantic information from a given image. Although the notion of capturing varying levels of detail (scale) is a common thread, they address the problem of accurately detecting varying-sized features from a given image, whereas we capture the evolution in symbolic knowledge about the environment.
\smallskip

\noindent\textit{{{B. Temporal Logic-Based Reactive Control Synthesis:}}}\\
 Many existing approaches for reactive control given temporal logic specifications assume accurate symbolic perception input~\cite{vasile2020reactive, ulusoy2014receding, kress2009temporal, kress2011mitigating, liu2013synthesis}.
{\dishalatest \cite{vasile2020reactive} and \cite{ulusoy2014receding} present sampling-based and automata-based reactive planning approaches, respectively, using potential functions to react to local requests while ensuring infinite-time satisfaction of LTL tasks}. In \cite{fu2014abstractions}, the authors propose abstraction-refinement approach for planning with incomplete sensor information given LTL specifications. On the contrary, we consider the problem of reactive planning where the symbolic sensing resolution incrementally improves (refines).
Assume-guarantee specifications reduce the otherwise computationally intensive, exponential-time synthesis problem as with LTL to a polynomial-time synthesis problem.
\cite{kress2011mitigating, kress2009temporal, liu2013synthesis} solve a game between sensor propositions (environment) and robot propositions (control actions) for synthesizing reactive control in abstract dynamic environments. Our work leverages this specification formalism for control synthesis. 

 Another line of research investigates safe decision-making under uncertainty\cite{kantaros2020reactive, lindemann2021reactive}. Generally, the problem results in a Partially Observable Markov Decision Process solved using approximate methods \cite{8248668, brechtel2014probabilistic, haesaert2019temporal, nilsson2018toward}. 
\cite{fu2016optimal} considers probabilistic semantic maps for generating optimal control plans.   
\cite{kantaros2022perception} proposes a sampling-based planning approach over an environment with semantic, metric uncertainties. {\dishalatest{Although these works are closely related, they are not reactive.}} Using Probabilistic STL,
\cite{da2021active} proposes a counter-example-guided approach under probabilistic perception uncertainty.  


The key difference is that we incorporate the \textit{refinement} in symbolic perception information into control synthesis. Our implementation considers symbolic perception input that is deterministic or is governed by a known probability distribution. However, the synthesis framework does not rely on quantification of environment uncertainty and is thus, independent of prior belief about the environment; investigating the incremental-resolution approach for a more general problem of reactive control synthesis in belief space is a topic for future research. 


{\disharevised 

}  

\smallskip 

\noindent \textit{{C. Contributions:}} \\This work is the first step in the direction of verified reactive planning with incremental perception. The main contributions of this work are as follows: 
1. We propose the problem of safe, high-level reactive control with incremental symbolic perception for an autonomous car.
2. We abstract the perception module and establish a novel incremental-resolution representation for the available symbolic perception information which facilitates the timely execution of the necessary behavior that the system must perform in order to satisfy rules of the road. 3. Additionally, we demonstrate with the help of case studies that this proposed approach helps synthesizing control inputs even in case of partial information about the environment. 

\section{Preliminaries: Models and Specifications} \label{sec:prelims}
\label{sec:preliminaries}
{\dishalatest The terms \textit{ego} and \textit{system} interchangeably refer to an autonomous robot car that performs the assigned {\disharevised temporal logic} tasks in a planar environment and operates using given control inputs. }
An \textit{environment} refers to everything external to the ego that can be sensed and cannot be controlled by the ego. {\disharevised For formal definitions, refer to Sec. \ref{sec:Problem}.} 



\begin{definition}[Deterministic Transition System, $\TS$]\label{def:TS}
A weighted deterministic transition system is a tuple $\TS = (X, x^\TS_0, \delta_\TS, \AP, h, w_\TS)$, where $X$ is a finite set of states; $x^\TS_0 \in X$ is the initial state; $\delta_\TS \subseteq X \times X$ is a set of transitions; $\AP$ is a set of properties (atomic propositions); $h : X \to 2^{\AP}$ is a labeling function; and $w_\TS : \delta_\TS \to \mathbb{R}^+$ is a positive weight function.
    
\end{definition}
We also denote a transition $(x, x') \in \delta_\TS$ by $x \to_\TS x'$. 
A finite or infinite state \textit{trajectory} $\mathbf{x} = x_0 x_1 \ldots$  such that 
$x_k \to_\TS x_{k+1}$ $ \forall k \geq 0$ generates an \emph{output trajectory} $\mathbf{o} = o_0 o_1 \ldots$, where $o_k = h(x_k) \forall k \geq 0$.
Furthermore, we denote the total weight of a finite run $\mathbf{x}$ as $\omega(\mathbf{x}) = \sum_{k=0}^{|\mathbf{x}|-1} \omega((x_k, x_{k+1}))$, where $|\mathbf{x}|$ is the length of $\mathbf{x}$.
We denote by $x \to^*_\TS x'$ the path in $\TS$ between $x$ and $x'$ with minimum weight.
The weight function $\omega$ may capture travel duration between states, distance, control effort, etc.

\smallskip

\noindent \textbf{\textit{Linear Temporal Logic (LTL) and Generalized Reactivity:}}
\label{sec:ltl}
Using LTL~\cite{baier2008principles}, specifications are expressed 
over a finite set of atomic propositions $p_i\in \AP$, $i=1,\ldots,|\AP|$.
Any LTL formula $\psi$ is built recursively via the syntax 
\begin{equation}
\vspace{-2pt}
\label{eq:syntax}
\psi ::= {\tru}\mid p_i \mid \lnot \psi \mid \psi \land \psi \mid \nex \psi \mid \psi \until \psi,
\vspace{-1pt}
\end{equation}



\noindent {\disharevised where, $\tru$, $\mathsf{false}$ are Boolean constants; \textit{negation} ($\neg$), \textit{conjunction} ($\land$) are Boolean operators; and \textit{next} ($\nex$) and \textit{until} ($\until$) denote the temporal operators. Additional operators allowed by LTL such as \textit{disjunction} ($\lor$), \textit{eventually} ($\eventually$), \textit{implication} ($\implies$), \textit{always} ($\always$), etc. can be derived using the operators defined in Eq.~\ref{eq:syntax}. 
%
%
For a detailed exposition to syntax and semantics of LTL, we refer the reader to~\cite{baier2008principles}.


In this work, we consider a special class of LTL formulas known as assume-guarantee specifications or GR(1) \cite{bloem2012synthesis}.  
Let $\Env$ denote the set of variables controlled by the environment and
$\Sys$ denote the set of variables controlled by the system.
The synthesis problem is a dynamic game between the uncontrollable environment and the to-be-synthesized system.
At each step of the game, the environment can choose the (Boolean) valuation 
of its variables in $\Env$ before the system can choose the {\disharevised Boolean} valuation 
of the variables in $\Sys$.

  
The general form of GR(1) formulae is
\begin{equation}
\label{eq:gr1}
\begin{aligned}
\textstyle
&\textstyle\psi:=\overbrace{\textstyle\l(\Theta^{\mathsf{e}}\land \always \psi^{\mathsf{e,safe}}\land \bigwedge_{k}^{K_{\mathsf{e}}}\always \eventually \psi^{\mathsf{e}}_k  \r)}^{\mbox{\footnotesize assumption on environment}} \quad \\ &\textstyle\hspace{2cm}\implies^{{sr}^{\footnotemark{}}}
 \underbrace{\l(\textstyle\Theta^{\mathsf{s}}\land \always \psi^{\mathsf{s,safe}} \land \bigwedge_{k}^{K_{\mathsf{s}}}\always \eventually \psi^{\mathsf{s}}_k   \r)}_{\mbox{\footnotesize guarantee on system}}
\end{aligned}
\end{equation} \footnotetext{\disharevised $\rightarrow^{sr}$ The subscript $sr$ denotes a strict realization of this formula, as defined in \cite{bloem2012synthesis}, for synthesis. This means that, to force the violation of the assumptions on the environment, the system may not precede with a violation of its own guarantees.  In the non-strict sense, this would lead to winning a game.}
\noindent $\Theta^{\mathsf{e}}$,$\Theta^{\mathsf{s}}$, $\psi^{\mathsf{e,safe}}$, $\psi^{\mathsf{s,safe}}$, $\psi^{\mathsf{e}}_k$ for $k \in K_{\mathsf{e}}$ and $\psi^{\mathsf{s}}_k$ for $k \in K_{\mathsf{s}}$ are  sub-formulae.
Precisely, starting from atomic propositions based on the valuations of $\Env$ and $\Sys$, the formulae $\Theta^{\mathsf{e}}$ and $\Theta^{\mathsf{s}}$ are composed via only Boolean operations 
($\land$,$\lor$, $\lnot$). {\dishalatest As such, }they restrict the allowed initial states for
the environment and system, respectively.
The safety formulae define the allowed transitions. $\psi^{\mathsf{e,safe}}$ is a Boolean formula over $\Env\cup\Sys\cup \nex \Env$ restricting the transitions of the environment, whereas $\psi^{\mathsf{s,safe}}$ is a Boolean formula over $\Env\cup\Sys\cup \nex \Env\cup \nex \Sys$. \emph{Progress} or \emph{liveness} are captured by the property $\always\eventually \psi$
which indicates that $\psi$ should happen infinitely often.
The progress goals for the environment $ \psi^{\mathsf{e}}_k$ and for the system  $ \psi^{\mathsf{s}}_k $ are Boolean formulae over $\Env\cup\Sys$.

%
 
\vspace{3 pt}
\section{Problem Formulation}
\label{sec:Problem}
\vspace{2 pt}

\subsection{Perception Model}
\label{subsec:env}



The signs and markings in the environment are detected by the ego which is assumed to be equipped with the necessary sensors, possibly with limited range and resolution.

In addition to the traditional detection of elements of the environment (e.g., stop\_sign, traffic\_light),
we assume that the perception module is capable of inferring multiple levels of
{\disharevised detail of the given element (e.g., sign\_color, sign\_type), see Fig.~\ref{fig:roadsignrefinement}. }
Thus, $\Env = \{x_1, \ldots, x_n\}$ consists of variables corresponding to varying levels of detail about the environment. These environment variables are atomic propositions and a particular variable is set to true when the event associated with it is detected in the environment.
In this work, we limit the environment variables to traffic signs, intersections and traffic lights. However, the proposed approach can be used in other scenarios wherein utilizing the partial knowledge about the environment may lead to safer control strategies e.g., search and rescue, exploration missions.

\begin{figure}[htb]
	\centering
\resizebox{0.55\linewidth}{!}{
\begin{tikzpicture}[node distance=2cm and 3cm]
	\node[inner sep=0pt] (Warning) {\includegraphics[width=1cm]{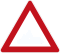}};;
	\node[below left of=Warning, inner sep=0pt] (WarningMinor)
	{ \includegraphics[width=1cm]{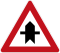}};
		\node[below right of=Warning, inner sep=0pt] (WarningLight)
	{\includegraphics[width=1cm]{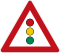}};
	\node[below  of=Warning, inner sep=0pt] (Warningpriority){\includegraphics[width=1cm]{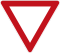}};
	\path[draw,->] (Warning) edge(WarningMinor);
	\path[draw,->] (Warning) edge(WarningLight);
		\path[draw,->] (Warning) edge(Warningpriority);
		
	\node[inner sep=0pt, right of =Warning,xshift=3cm] (reg) {{\includegraphics[width=1cm]{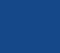}}};
		\node[inner sep=0pt, below right of =reg] (regped) {\includegraphics[width=1cm]{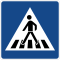}};
    	\node[inner sep=0pt, below left of =reg] (regzone) {\includegraphics[width=1cm]{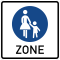}};
\path[draw,->](reg)->(regped);
\path[draw,->](reg)->(regzone);

\node[ above right of=Warning, inner sep=0pt,xshift=1cm ] (unknown) {\includegraphics[width=1cm]{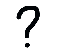}};
\path[draw,->](unknown)->(Warning);
\path[draw,->](unknown)->(reg);
\end{tikzpicture}
}
\caption{Refinement tree of ego car's perception. }
\label{fig:roadsignrefinement}
\vspace{-7pt}
\end{figure}
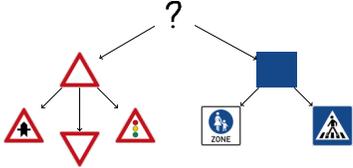

\subsection{Robot Model}

Consider an ego car deployed in a planar workspace $\mathcal{E} \subseteq \mathbb{R}^2$. We assume its dynamics to be governed by $\dot{x} = f(x,u)$, where $x \in \mathbb{X}$ is the state space and $u \in \mathbb{U}$ represents the control space. The state of the ego may capture its location, orientation as well as velocity. We abstract ego's motion in the workspace as a deterministic transition system $\TS$. 



The control space of ego is abstracted via a finite {\disharevised set} of \textit{system variables} $\Sys = \{y_1, y_2, \ldots, y_n\}$
denoting actions such as $stop$, $move\_slowly$, etc. {\disharevised These system variables are atomic propositions. A system variable is set to true when the corresponding action is being executed by the robot.} 
{The constraints between actions are captured in the system guarantees LTL property,} e.g.,
$come\_to\_stop \implies \lnot move$.

We consider a receding horizon approach for navigating the environment where a nominal route is assumed to be available to the ego. 
The ego follows a time-varying goal with respect to an abstract symbolic reference frame centered at the ego. Here, the goal refers to the next way-point to be reached while making progress on the given nominal route as expressed by the GR(1) formulae.


\subsection{Rules}
Intuitively, a rule associated with a specific environment configuration can be defined as the \textit{expected behavior} by the ego upon encountering that configuration. 
Formally, a set of rules $\phi$ is a set of GR(1) formulae of the form~\eqref{eq:gr1} that establish the relation between the perceived environment state and the expected robot behavior, e.g., $\lnot sign \Rightarrow \eventually \mathit{max\_speed}$.

{\noindent \textbf{\textit{Example}}:
Consider an ego approaching a work zone. Let $\Env = \{work\_zone\}$ and $\Sys = \break \{move\_slow\}$. The rule to be satisfied is ``When work zone is encountered, slow down to the posted speed limit; If no work zone, move with maximum allowed speed". 

\begin{table}[H]
\vspace{-3 pt}
\centering
\begin{tabular}{p{0.15\linewidth}p{0.37\linewidth}p{0.33\linewidth}}
\hline
 Conditions & Environment & System\\
\hline
 Initial  & $\Theta^e = \lnot work\_zone,$ & $\Theta^s = \lnot move\_slow $\\
 Safety  & $\psi^{\mathsf{e, safe}} = True$ 
 (environments with / without work zones are admissible) 
& $\psi^{\mathsf{s,safe}} = \Next(work\_zone) \implies \Next(move\_slow)$\\
 Progress  & $\psi^{\mathsf{e}} = \lnot work\_zone$ & $\psi^{\mathsf{s}} = \lnot move\_slow$.
\end{tabular}
\vspace{-5pt}
\end{table}
}

\noindent \textit{Remark}: Multiple environment and system variables can be true at the same time. Furthermore, if two or more system variables cannot be true simultaneously, it should be explicitly specified as a system safety condition.


\begin{problem}[Hierarchical Perception-based Verified Reactive Planning]
\label{pb:problem-control-synth}
Given robot model, set of environment variables $\Env$, system variables $\Sys$, and the rules of the road $\phi$,  synthesize a symbolic controller (if feasible) such that the generated behavior plan for the robot to follow the nominal route
satisfies all rules $\phi$ for all valid valuations of the elements of $\Env$.
\end{problem}





\section{Solution}
\label{sec:solution}

This section presents a solution to Problem \ref{pb:problem-control-synth}. 
In order to synthesize a winning strategy for the controllable system and uncontrollable environment, we use a receding horizon control synthesis approach.
The abstractions of perception and control are centered at the ego vehicle.
Thus, these abstractions are updated at each time step as the ego exhibits different control actions.
The abstract models along with the rules of the road are internally converted into a GR(1) specification and then into a GR(1) game by a control synthesis tool. 
With the help of GR(1) specifications, we capture both - the hierarchy in the environment variables and ego's motion.
The control strategies for any valid environment states are obtained as a solution to the GR(1) game (if feasible). {\dishalatest Addressing reactivity at the specification and control level \cite{kress2011mitigating, belta2007symbolic} we assume that robust low-level controllers are available that handle the implementation level uncertainties such as obstacle avoidance, interactions with pedestrians.}
The detailed solution is as follows:

\subsection{Incremental Symbolic Model of Perception}
The incremental symbolic knowledge about the environment is represented using a directed tree structure referred to as the \textit{symbolic refinement tree} defined as follows: 

\begin{definition}[Symbolic Refinement Tree for Perception]
\label{def:symbolic-refinement-tree}
The semantic relationship between symbols is captured by a directed tree $\mathcal{G} = (\Env, \Delta)$ where $\Delta$ establishes how the nodes in $\Env$ are related to each other.
The root of the node corresponds to whether there is an object in the environment.
Each subsequent layer of the tree adds more information, capturing a refinement of the object's meaning (semantics).
\end{definition}

\begin{definition}[Ground Variables]
\label{def:base-vars}
The set $\Env^{gr} \subseteq \Env$ is a set of all environment variables that correspond to the terminal nodes of the perception tree. These variables correspond to the ground truth of perception, e.g., \textit{stop\_sign}, \textit{crosswalk\_sign}.
\end{definition}

\begin{definition}[Derived Variables]
\label{def:dev-vars}
The set $\Env^{dr} \subseteq \Env$ consists of all other environment variables that lead up to the ground variables.
These variables correspond to the progress of robot's knowledge towards the ground truth of perception $\Env^{gr}$ thereby capturing the symbolic multi-resolution capabilities of the perception module, e.g., \textit{sign\_shape}, \textit{mandatory\_sign}, \textit{danger\_warning\_sign}. \end{definition}
\vspace{-2pt}
Naturally, we have $\Env^{gr} \cap \Env^{dr} = \emptyset$ and $\Env^{gr} \cup \Env^{dr} = \Env$. 
Detection results in setting one or multiple environmental variables $\Env$ to true. In what follows, this occurrence of true environment variables is referred to as an \textit{event}.

{\dishalatest \noindent \textit{Remark:} We assume that detection is persistent, i.e., {coupled detection and tracking methods within the perception module. Consequently, once a variable is set, it will remain set or set a refined version of it from its descendants
in $\mathcal{G}$.}}

{\dishalatest With respect to the abstract symbolic reference of the robot, the subset of $X_c$ over which the ground and derived variables can be detected is referred to as \textit{perception horizon.}}

%
%
The refinement tree $\mathcal{G}$ is converted into GR(1) formulae in assume-guarantee form, and provided to Tulip\cite{wongpiromsarn2011tulip} for synthesis.
The formulae capture both the refinement and persistence of information, see Sec.~\ref{sec:case study} for examples.

\subsection{Robot Motion Planning}
A \emph{cell decomposition} of $\mathbb{X}$ is a set $C$ of subsets of $\mathbb{X}$ such that $\bigcup_{A\in C} A = \mathbb{X}$, $cl(A)=A$ for all $A\in C$, where $int(A)$ and $cl(A)$ denote the interior and closure of set $A$, respectively. A cell decomposition of $\mathbb{X}$ is an \emph{exact cover} of $\mathbb{X}$
with closed cells that share only boundaries if $int(A)\cap int(A') = \emptyset$ for all $A, A' \in C$.


\smallskip
\noindent{\bfseries Finite state abstraction.}
 Let $\TS_C$ be a transition system representing the robot model abstraction such that
the state space $X_C$ corresponds to a cell decomposition of $\mathbb{X}$.
A transition between two cells $c$ and $c'$, $c, c' \in X_c$ exists if there is $\xi \in c$, $\xi' \in c'$, and  there exists a sequence of high-level control actions $(y_a,\ldots, y_b)$ such that 
their execution results in driving the system from $\xi$ to $\xi'$.

\begin{definition}[Target Cell]
Given $X_c$, the target cell $c_{target}$ is the cell in the perception horizon {\dishalatest sensing range} of the system where a base variable $x^{gr} \in \Env^{gr}$ (see~\ref{subsec:env}) is present. 
\end{definition}

\noindent{\bfseries Finite movement abstraction.}
Consider the transition system $\TS_T$  that consists of a state space $X_T\subset X_C\times X_C$ such that $(c_1,c_2)\in X_T$ if there exists a transition in $\TS_C$ between the cells $c_1$ and $c_2$. 
We can now label the states based on both the location of the car and the transition that the car is making. The latter allows us to associate {\dishalatest propositions} such as  \emph{moving} or \emph{\disharevised stationary} to the car. This also implies whether new information will enter the perception model.

\subsection{Control Synthesis as a GR(1) Game}
\label{subsec: gr1}
Both the incremental symbolic model of the perception and the symbolic model of the motion planning problem can now be captured together with the rules into one GR(1) game.
For control synthesis, we use the Omega solver \cite{filippidis2016symbolic}
in Tulip~\cite{wongpiromsarn2011tulip,filippidis2016control},  which is an abstraction-based control synthesis tool.
Internally, it constructs an enumerated transducer based on the given GR(1) specification and solves a Streett game
-- here solving refers to finding a controller that ensures the satisfaction of GR(1) formula for any admissible behavior of the uncontrolled {\disharevised environment}.
It is important to note that GR(1) can only be reactive to ``known" unknowns, i.e.,
if the environment variable takes a value other than the imposed assumptions, no guarantees can be provided on the system behavior.
Moreover, due to the particular structure requirement of GR(1), some temporal formulae like $\eventually \always \psi$ cannot be directly expressed as opposed to \cite{wolff2013efficient}.
However, the GR(1) structure is sufficiently expressive for the problem being addressed \cite{kress2009temporal},\cite{bloem2012synthesis} while also avoiding the exponential time complexity as with LTL. A detailed description of GR(1) synthesis can be found in~\cite{bloem2012synthesis},~\cite{Piterman2006}. 

The continuous execution of the discrete solution based on the abstract models can be achieved in a provably correct fashion if there exists a similarity relation for the abstractions of continuous models~\cite{871304}. For each system variable $y_i$, we associate a continuous controller $\Lambda(y_i)$.
Such low-level controllers can be computed using standard methods~\cite{LinLav09} and are assumed to be available.
If the a system variable $y_i$ is set to \emph{True}, the ego executes control $\Lambda(y_i)$. 




\section{Resolving traffic regulations with temporal logic control: Case {\disharevised Studies}}
\label{sec:case study}

Given limited perception, this section demonstrates how the available information can be utilized by the ego to ensure continued conformance to traffic rules. 
In both cases, the ego is assumed to be equipped with necessary sensors as well as recognition and tracking algorithms that provide symbolic perception input with respect to the ego's frame. The ego follows a nominal route defined in ego's reference frame where at least the immediate way-point is always within the perception horizon. The presence or absence of signs as well as their exact location are not known a priori.
{\disharevised \subsection{Case Study 1}}
We first consider a fully observable environment wherein the ego is moving on a straight road section with one traffic sign. The ego is provided with 
necessary way points. The upcoming road of the ego vehicle can be partitioned into cells as depicted in Fig.~\ref{fig:stop_refinement}.
{\disharevised Depending on the distance from $c_{target}=1$, we obtain} refinement of the road sign information while the car is approaching the road sign.
The levels of knowledge of the ego about the environment are \{{\em sign\_present, sign\_type, sign\_shape, exact\_sign}\}. These will serve as environment variables in the 2-player game as described in Section \ref{subsec: gr1} and form the nodes of $\mathcal{G}$. Initially, no signs are present. The condition for environment progress is ``infinitely often, no signs" and is formally given by
    $\Box \eventually \neg \mathit{\disha exact\_sign.}$

\begin{figure}[htp]
\vspace{-10pt}
	\centering
	\includegraphics[trim={2.4cm 4.8cm 0.9cm 2.5cm}, clip, scale=0.32 ]{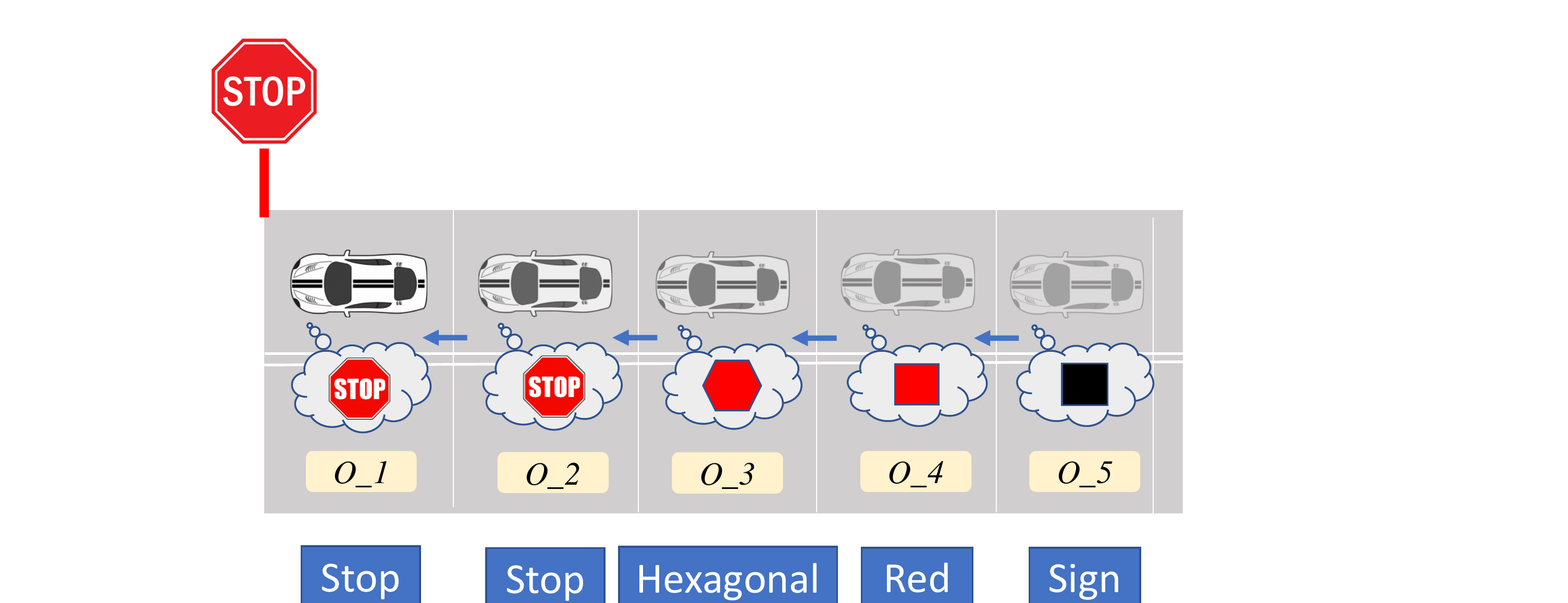}
	\vspace{-2 pt}
	\caption[Example of road environment]{The knowledge of the ego car about the environment increases as it moves closer to the stop sign.}
	\label{fig:stop_refinement}
\end{figure}

\newcommand{\emptys}{\tikz{\node[rectangle, draw,inner sep=5pt] {};}}
\newcommand{\obj}{\tikz{\node[rectangle, draw,inner sep=5pt, fill=black] {};}} \newcommand{\warn}{\includegraphics[width=0.4cm]{DE/warning/empty_warning}} \newcommand{\minor}{\includegraphics[width=0.4cm]{DE/warning/crossroads_with_a_minor_road}} \newcommand{\tsign}{\includegraphics[width=0.4cm]{DE/warning/traffic_signals_ahead}}
\newcommand{\yield}{\includegraphics[width=0.4cm]{DE/priority/yield}}
\newcommand{\bsign}{\includegraphics[width=0.4cm]{DE/special_regulations/Blue}}
\newcommand{\peds}{\includegraphics[width=0.4cm]{DE/special_regulations/pedestrian_crossing}}
\newcommand{\pedz}{\includegraphics[width=0.4cm]{DE/special_regulations/pedestrian_zone}}
The perception horizon is up to 5 cells in front of ego, variables $o_i$,  $i=1,\ldots, 5$ represent the identified {{symbols}} in these cells in Fig.~\ref{fig:stop_refinement}.
The stop sign is present in the last cell of discretization i.e. in $o_1$. 
The knowledge about the environment evolves as follows: At the first level of the refinement tree, when the stop sign is farther away, the car can only detect the presence (black box) or absence (white box) of a sign, $o_5$ = $\{$\tikz{\node[rectangle, draw,inner sep=5pt, fill=black] {};}, \tikz{\node[rectangle, draw,inner sep=5pt] {};}$\}$. At 
 the following level, the color of the sign can be identified $o_4 \in  \{\includegraphics[width=0.4cm]{DE/special_regulations/Blue}, \tikz{\node[rectangle, draw,inner sep=5pt, fill=yellow] {};}, \tikz{\node[rectangle, draw,inner sep=5pt, fill=red] {};}\}$ and thus, whether it is a regulatory/warning/priority sign, etc. Next, the shape of the sign is detected $o_3 \in \{\tikz{\node[regular polygon, regular polygon sides=6, fill=red] {};},  \includegraphics[width=0.4cm]{DE/priority/yield}, \tikz{\node[circle, draw=black, fill=red] {};}, \tikz{\node[diamond, draw=black, inner sep=3pt, fill=yellow] {};}, \includegraphics[width=0.4cm]{DE/special_regulations/Blue}\}$ followed by the exact sign at the last level: $o_2, o_1 \in \{$  \includegraphics[width=0.4cm]{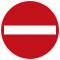}, 
 \includegraphics[width=0.4cm]{DE/warning/traffic_signals_ahead}, 
 \includegraphics[width=0.4cm]{DE/priority/yield}, %
 \includegraphics[width=0.4cm]{DE/special_regulations/pedestrian_crossing}, %
 \includegraphics[width=0.4cm]{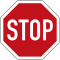}, \includegraphics[width=0.4cm]{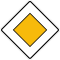}  $\}$.
 
The environment variables evolve as follows as the car travels from cell $c_5$ to $c_1$-
1) If no sign in cell $c_i$: in the next step, no sign in the current cell $c_{i-1}$ (eq. ~\eqref{eq:perception-no-sign}-\eqref{eq:perception-some-sign}),
2) if sign in $c_i$: in the next step, we get more information about it in cell $c_{i-1}$, (eq. \eqref{eq:perception-sign-color}-\eqref{eq:perception-regulatory-sign}), and, thus, we move to the next level of the perception refinement tree.
\begin{align}
 &(o_i = \emptys)\rightarrow \bigcirc(o_{i-1}=\emptys) \label{eq:perception-no-sign},\\
 &(o_5 = \obj)\rightarrow \bigcirc(o_{4}=\obj) \label{eq:perception-some-sign},\\
 &(o_4 = \obj)\rightarrow \bigcirc(o_{3}=\bsign\vee o_{3}=\tikz{\node[rectangle, draw=black,inner sep=5pt, fill=red] {};}, \vee o_{3}=\tikz{\node[rectangle, draw=black,inner sep=5pt, fill=yellow] {};}) \label{eq:perception-sign-color},\\
 &\vdots\\
  & ( o_{2}=\bsign)\rightarrow \bigcirc  (o_{1}=\peds\vee o_{1}=\pedz). \label{eq:perception-regulatory-sign}
\end{align}

Fig.~\ref{fig:stop_refinement} shows the evolution of knowledge of the system and Fig. \ref{fig:stop_tree} depicts the interaction between perception and control abstraction for a `STOP' sign. In this case, the environment variables are  \{{\em sign\_present, sign\_red, sign\_hexagonal, stop\_sign}\} and the environment progress condition is $\Box \eventually \neg \mathit{\disha stop\_sign}$. {\disharevised The control inputs for the system} are : \{{\em move, attention, slow\_down, prepare\_to\_stop, stop}\}. The system progress rule is given by $\Box \eventually \mathit{move}$. The system safety rule is $\psi^{\mathsf{s,safe}} = (\textit{stop\_sign} \implies \textit{stop} \land \bigcirc(move))$
. The control input for the system progresses towards making a complete stop as the knowledge about the environment increases (see Fig.~\ref{fig:stop_tree}). This points at the natural interrelation between perception and control abstraction.

\begin{figure}[htp]
	\centering
	\includegraphics[ scale=0.32, trim = 3.5cm 0.5cm 2cm 0.8cm, clip]{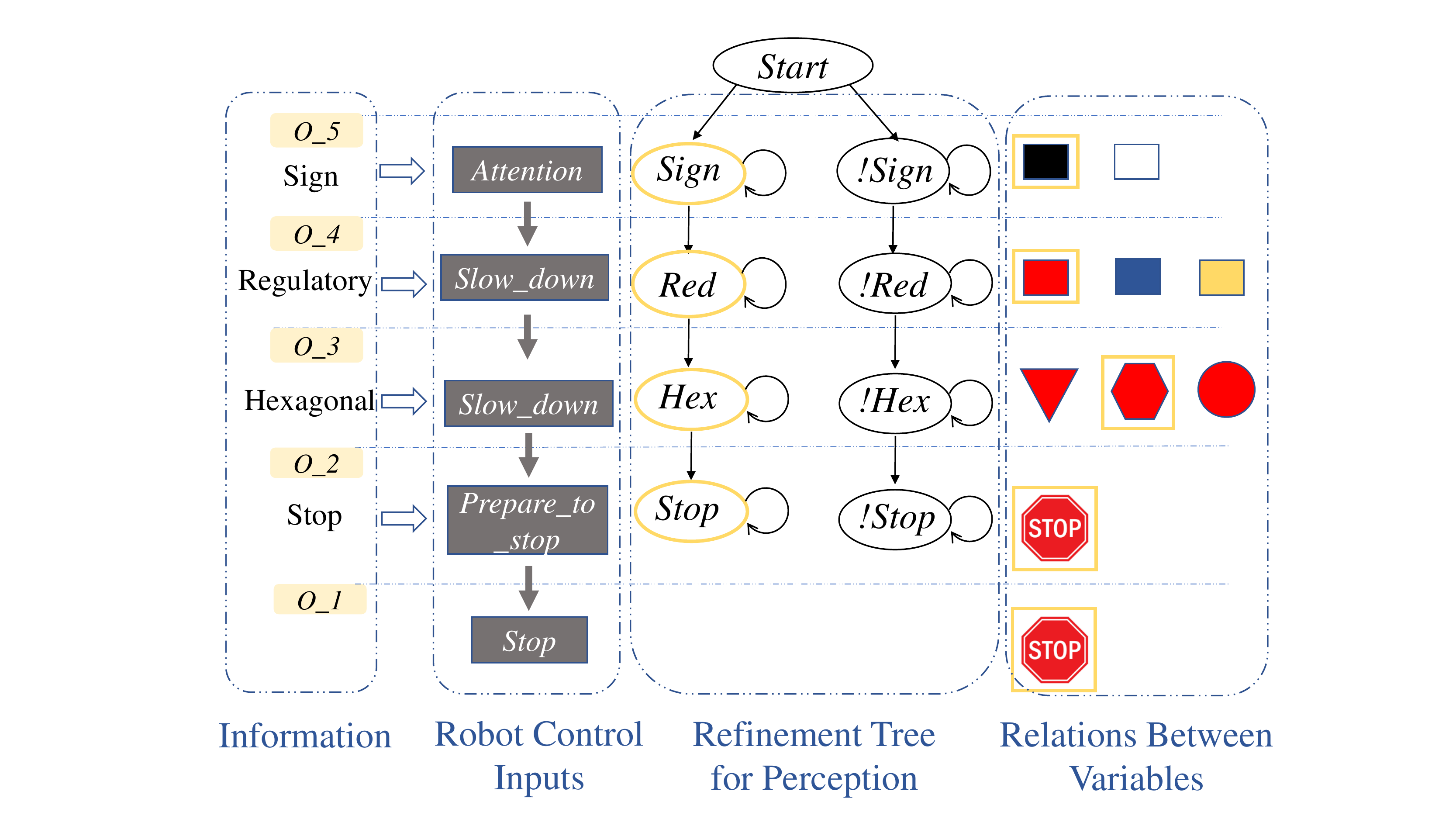}\caption[The interplay between perception and control modules]{ The interplay between perception and control modules: As the robot progresses towards $c_{target}$, its knowledge about the environment incrementally improves, resulting in the appropriate control actions.}
	\label{fig:stop_tree}
\end{figure}

\vspace{-3 pt}

\subsection{Case Study II}

\begin{figure}[htp]
\flushright
	\centering
 \vspace{-2pt}
	\includegraphics[scale=0.3, trim = 1cm 1cm 2cm 0.5cm,clip]{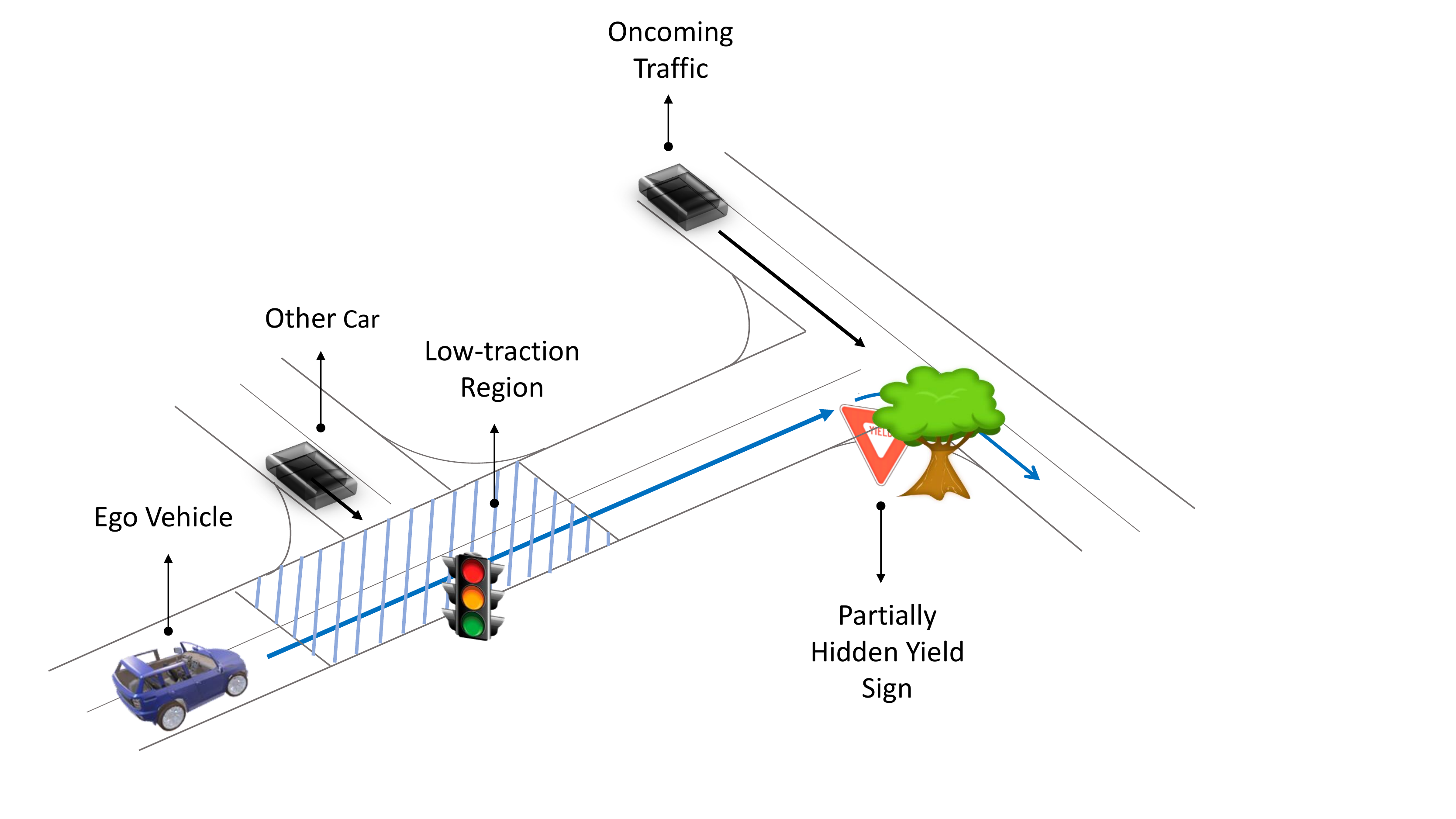}
	\caption[The interplay between perception and control modules]{The ego car approaching a traffic signal in a reduced traction region followed by a partially visible yield sign. In both cases the car needs to react to partial information to  satisfy the rules.  }
	\label{fig:roadcasestudy}
\end{figure}

Extending the previous case study, we now consider the perception knowledge to be probabilistic pertaining to the uncertainties associated with partially occluded traffic lights and signs, sensor limitations, different weather and road conditions, inaccuracies associated with detection, tracking, etc. We compare our proposed framework with the classical perception-planning approach  (\textit{baseline}) that does not consider symbolic refinement in perception. Note that, we employ the same high-level reactive control synthesis module for both cases and the only difference lies in whether or not incremental symbolic perception information is utilized for control synthesis. {\dishalatest Moreover, the probabilities are used only for simulating the environment. The control synthesis is independent of these beliefs.} 

Fig.~\ref{fig:roadcasestudy} shows the test scenario where the ego car first encounters a traffic light at an intersection in a \textit{reduced traction} region followed by a partially visible \textit{yield} sign.

The probabilistic perception refinement is obtained as follows: The probability distribution over the terminal nodes $\chi^{gr}$ indicates the probability of encountering a particular sign in the environment and is obtained by the state-of-the-art traffic sign detection algorithms e.g., \cite{liang2020traffic, laguna2014traffic}. The probability of each derived variable $x_{dr} \in \chi^{dr}$ is then obtained using the total probability rule used recursively over the refinement tree.
For simulating the baseline, we only have access to the probability distribution over the terminal nodes. For both cases, the identified environment variable is then used for control synthesis. We study the reactive control synthesis for following two events.


\subsubsection{\textit{Event: Traffic light}} Of particular interest are the cases where the traffic light is red and the ego car needs to decelerate from $v_{max}$ to 0 within the remaining distance to $c_{target}$ (the cell containing traffic light) while accounting for the reduced friction condition. 
    As the stopping distance is higher in this case, a safe behavior would imply starting to slow down as early as possible and avoiding hard stops.
    
\subsubsection{\textit{Event: Yield Sign}} Since the yield sign is partially visible, the perception module may not be able to accurately identify the sign in time for the system to be able to react and, thus, may lead to a collision with oncoming traffic.




\indent The environment and system variables for both cases are given in Table \ref{tab}. The system variables are of the form: \textit{action\_in\_x} where action refers to the control input to be executed and $x$ is the number of cells within which to execute this input. e.g., \textit{stop\_in\_4} implies come to a complete stop in 4 cells. \textit{hard\_stop} and \textit{yield\_now} indicate that the action needs to be executed within one cell and \textit{infeasible} indicates that the system has already reached $c_{target}$ without detecting the event. Fig. \ref{fig:case_study2}
shows the comparison between the actions chosen by the system in response to the environment. 

\begin{table}[t]
\caption{Environment and System Variables}
\centering
\begin{tabular}{p{0.09\linewidth}p{0.32\linewidth}p{0.45\linewidth}}
\hline
Event & Baseline & Our System\\
\hline
 \multirow {1}{*}  Traffic Light & \textbf{env\_vars} = \{\textit{light\_color, reduced\_traction}\} &  \textbf{env\_vars }= \{\textit{intersection, traffic\_light, light\_color, reduced\_traction}\}\\
 &\textbf{sys\_vars} = \{\textit{stop\_in\_4, stop\_in\_3, stop\_in\_2, hard\_stop, infeasible}\} &\textbf{sys\_vars} = \{\textit{stop\_in\_4, stop\_in\_3, stop\_in\_2, hard\_stop, infeasible}\}\\
\hline
 \multirow {1}{*} Yield Sign & \textbf{env\_vars} = \{\textit{exact\_sign}\} &  \textbf{env\_vars} =  \{{\em sign\_present, sign\_type, sign\_shape, exact\_sign}\}\\
  &\textbf{sys\_vars} = \{\textit{yield\_in\_4, yield\_in\_3, yield\_in\_2, yield\_in\_1, infeasible}\} &\textbf{sys\_vars} = \{\textit{yield\_in\_4, yield\_in\_3, yield\_in\_2, yield\_in\_1, infeasible}\}\\
  \label{tab}
\end{tabular}
\vspace{-25pt}
\end{table}

{\textbf{ \textit{Performance.}}}
\label{sec:performance-measures}
We consider a distance-based metric defined over the robot motion model to measure the goodness of an already synthesized plan as follows: Let $c_d \in X_c$ be the cell where a derived variable $x_{dr}$ is first sensed in the given cell decomposition as defined in Sec.~\ref{sec:solution}.
The performance $\mathsf{s}_{yi}$ corresponding to a chosen control action $y_i$ by the system is defined as: 
{\dishalatest
    $\mathsf{s}_{yi} = \omega(c_d \to^*_{\TS_C} c_{target})$,
where $\omega(c_d \to^*_{\TS_C} c_{target})$ represents the topological distance between
$c_d$ and $c_{target}$ in $\TS_C$. Thus, the higher the distance available to react, the better is the performance.}

\begin{figure}[htb]
	\centering
 \vspace{-5pt}
 {\includegraphics[scale=0.25]{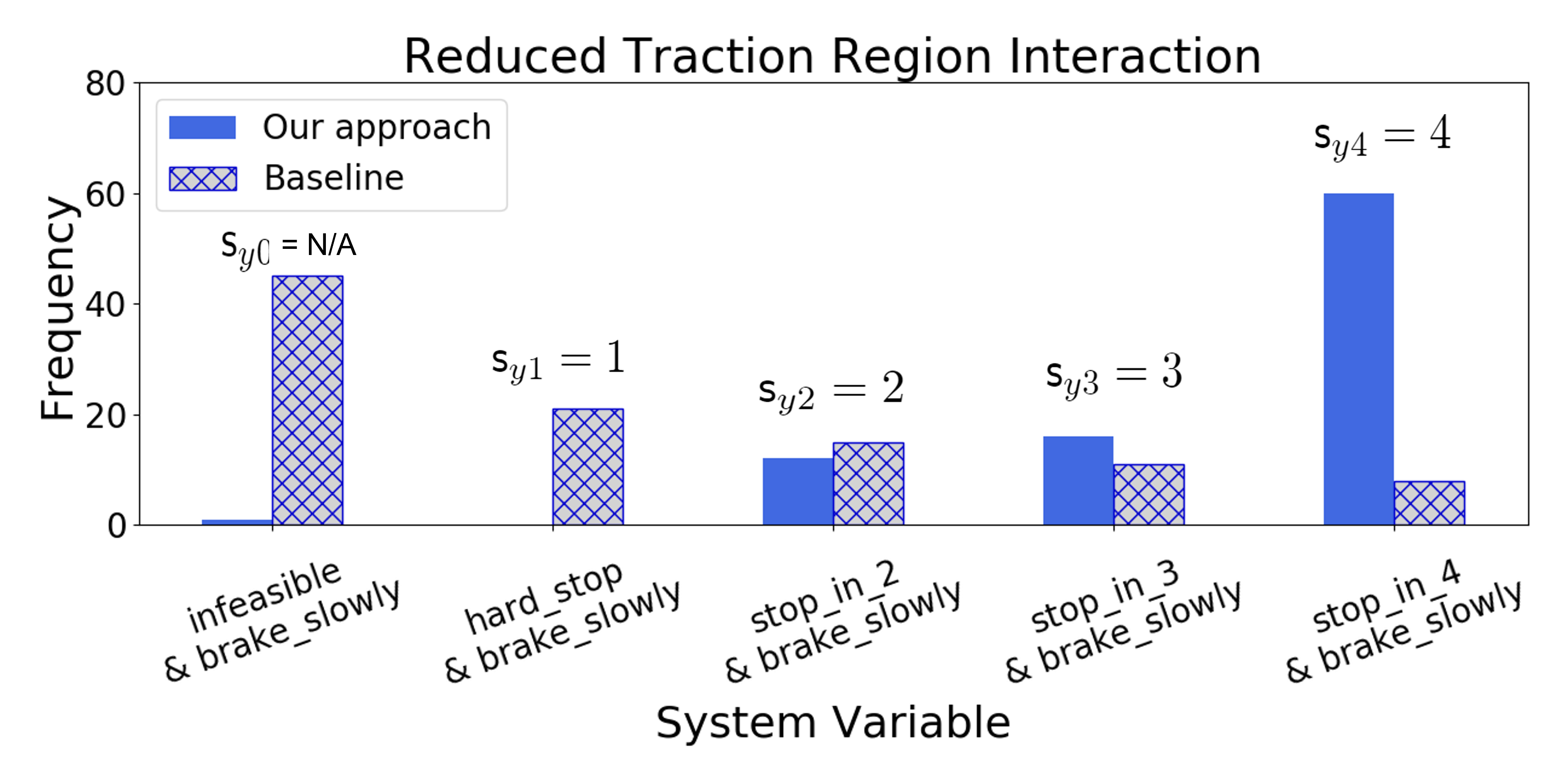}\label{fig:traction}}\\
	\centering
 {\includegraphics[scale=0.25, clip]{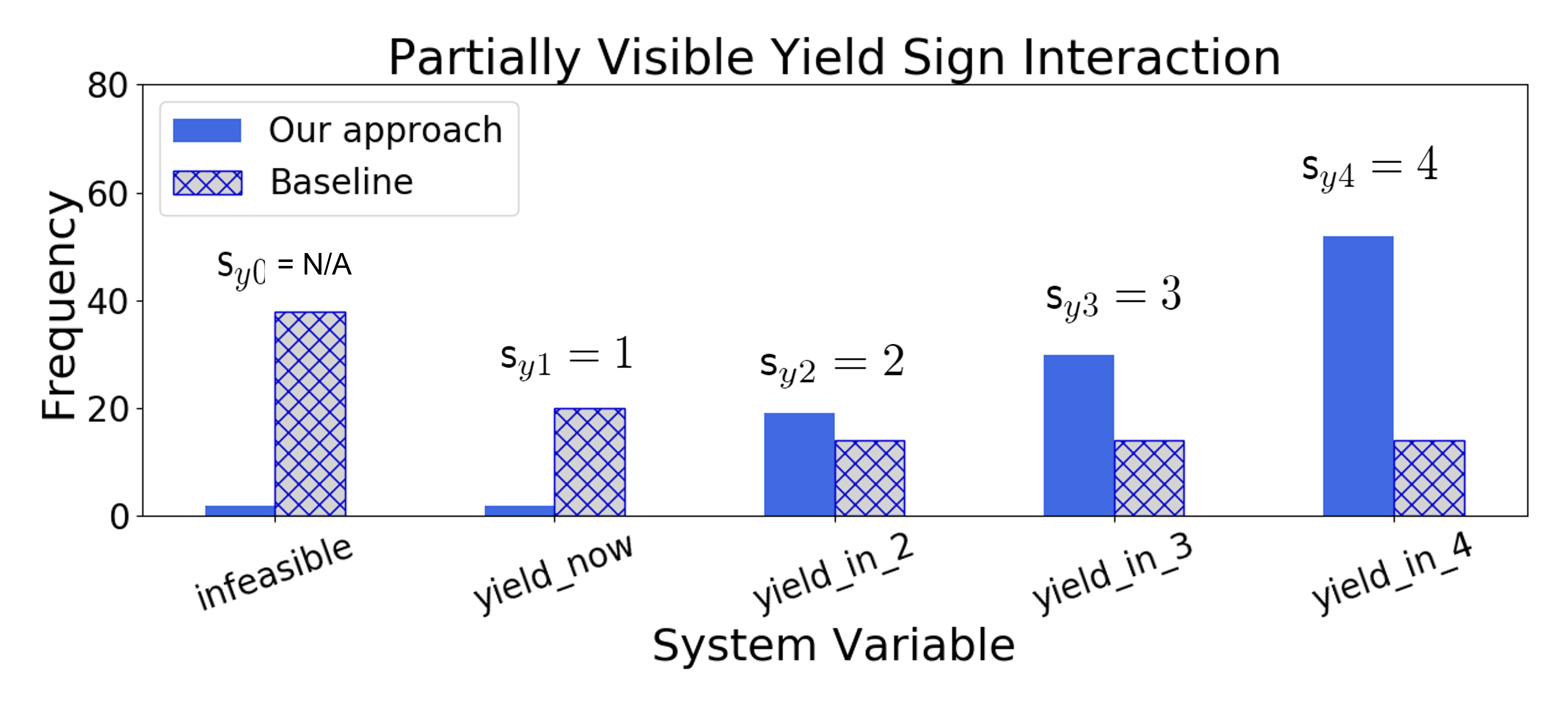}\label{fig:yield}}
 \vspace{-10pt}
	\caption{The comparison between control inputs obtained for a system with (\textit{blue}) and without (\textit{hatched grey}) incremental-resolution perception for interaction for 100 trials. Performance $\mathsf{s}$ increases from left to right as the available distance to react increases.}
	\label{fig:case_study2}
\end{figure}
{\dishalatest Fig.~\ref{fig:case_study2} shows the comparison between the control inputs for a system with and without incremental-resolution perception for n=100 trials. The frequency indicates the number of times a particular control action is chosen by the ego.{ Furthermore, the performance values of the chosen control action are indicated. For simplicity, we assume cells of unit length, thus if $c_d = c_2,$ naturally $\mathsf{s_{yi} = }$ 2 and so on. Since the distance to execute the chosen maneuver in the discretized environment is higher, the performance increases from left to right in Fig.~\ref{fig:case_study2}. 
Evidently, the ego consistently achieves higher performance when leveraging the symbolic relationship among input environment variables.
}}

{\section{\dishalatest Conclusion}}
\label{sec:conclusions}

This work is a step towards representing and utilizing incremental symbolic perception information for synthesizing verified reactive control plans for an autonomous robot car. We propose a novel incremental-resolution symbolic tree for perception - an abstraction that captures gradual improvements in the ego's knowledge about the environment as well as the symbolic relationships among various environment variables. Furthermore, we combine the abstractions of information gathering and motion control along with rules of the road to be satisfied in a single GR(1) game. The case studies demonstrate an improved performance when leveraging the proposed framework with respect to execution of high-level control actions. Future directions involve experimental validation and safe control synthesis under environment uncertainty with incremental-resolution perception. 
\newpage
\bibliographystyle{IEEEtran}
\bibliography{references}
\end{document}